\pdfoutput=1

\documentclass[11pt]{article}
\usepackage{amsmath}
\usepackage[final]{acl}

\usepackage{times}
\usepackage{latexsym}
\usepackage{pifont}
\usepackage{booktabs}
\usepackage{tabularx}
\usepackage[table]{xcolor}
\usepackage[ruled,vlined]{algorithm2e}

\usepackage[T1]{fontenc}

\usepackage[utf8]{inputenc}

\usepackage{microtype}

\usepackage{inconsolata}

\usepackage{graphicx}
\usepackage{multirow} 

\title{MI-PRUN: Optimize Large Language Model Pruning via Mutual Information}

\author{
\bf Hao Zhang$^{1,2,3}$\thanks{zhanghao233@mails.ucas.ac.cn},
Zhibin Zhang$^{1}$,
Guangxin Wu$^{1,2}$,
He Chen$^{1,2}$,
Jiafeng Guo$^{1}$,
Xueqi Cheng$^{1}$ \\
\normalsize{$^{1}$Institute of Computing Technology, Chinese Academy of Sciences} \\
\normalsize{$^{2}$University of Chinese Academy of Sciences} \\
\normalsize{$^{3}$School of Advanced Interdisciplinary Sciences, University of Chinese Academy of Sciences}
}

\begin{document}
\maketitle
\begin{abstract}
Large Language Models (LLMs) have become indispensable across various domains, but this comes at the cost of substantial computational and memory resources. Model pruning addresses this by removing redundant components from models. In particular, block pruning can achieve significant compression and inference acceleration. However, existing block pruning methods are often unstable and struggle to attain globally optimal solutions. In this paper, we propose a mutual information based pruning method MI-PRUN for LLMs. Specifically, we leverages mutual information to identify redundant blocks by evaluating transitions in hidden states. Additionally, we incorporate the Data Processing Inequality (DPI) to reveal the relationship between the importance of entire contiguous blocks and that of individual blocks. Moreover, we develop the Fast-Block-Select algorithm, which iteratively updates block combinations to achieve a globally optimal solution while significantly improving the efficiency. Extensive experiments across various models and datasets demonstrate the stability and effectiveness of our method.
\end{abstract}

\section{Introduction}

Large Language Models (LLMs) have demonstrated remarkable capabilities in natural language processing, enabling a wide range of applications such as question answering, summarization, and code generation \cite{ding2022, qin2023,zhu2023,li2023}. Moreover, these models also demonstrate exceptional performance across a wide range of other domains, including medicine \cite{qi2025mediaug,luo2025pathohr,cong2025hierarchical,qi2025medconv}, security \cite{
wu2025sugar}, and various social tasks \cite{zhang2025can,he2025enhancing}.  However, as the scale of models expands, the challenges faced in practical deployment also increase. The large size and computational requirements of the models lead to high memory costs and inference delays. In pursuit of lightweight deployment for LLMs, the research community has developed an array of model compression strategies, such as model pruning \cite{ma2023llm,ashkboos2024slicegpt,li2023communication,han2015deep}, quantization \cite{zhou2021smoothquant,cai2023gptq,zhou2024framequant} and knowledge distillation \cite{yang2021knowledge,2024CDL}. These approaches aim to lighten the computational load, enhancing deployability with constrained resources. Among them, model pruning is essential for enhancing the efficiency of deep learning models by eliminating redundant weights or neurons without compromising performance.

While the benefits of pruning are clear, applying it to LLMs still remains a formidable challenge \cite{ma2023llm,ashkboos2024slicegpt,li2023communication,han2015learning,fang2023depgraph}. Current pruning techniques often fail to deliver significant acceleration. To address this issue, inspired by the redundancy in the depth of LLMs, some studies have proposed pruning entire blocks rather than individual components to achieve compression \cite{men2024shortgpt,yang2024laco,chen2024streamlining,song2024sleb,kim2024shortened}. However, existing methods suffer from two main limitations: (1) They typically rely on metrics such as cosine similarity or perplexity. These metrics are highly dependent on the calibration set, and different calibration sets often yield inconsistent results, leading to instability. (2) Existing approaches often select pruned block combinations using greedy algorithms, which do not guarantee an optimal solution and may compromise performance.

Additionally, some studies \cite{fan2021layer,huang2024large,westphal2024mutual,ganesh2021mint,wu2026iterative} have explored pruning neurons or activations across layers by computing the mutual information between neurons. However, these methods only measure the mutual information between pairs of units, which is often insufficient because they ignore the collective dependencies among multiple units. Moreover, computing mutual information for each pair of neurons individually also incurs a significant computational cost.

In this paper, we propose a mutual information based pruning method MI-PRUN for LLMs \textbf{(note that we do not simply apply mutual information to block selection; our method overcomes the aforementioned limitations of mutual information based pruning)}. Specifically, we leverages mutual information to identify and remove redundant blocks by evaluating transitions in hidden states. Additionally, we incorporate the Data Processing Inequality (DPI) \cite{beaudry2011intuitive,merhav2012data,braverman2016communication} to reveal the relationship between the importance of entire contiguous blocks and that of individual blocks. Moreover, we employ an iterative block selection algorithm to continuously refine the combination of removed blocks. To further improve efficiency, we introduce the Fast-Block-Select algorithm, which uses heuristic strategies to quickly identify the most promising block removal. Extensive experiments demonstrate the stability and effectiveness of our method.

Our main contributions are as follows:
\begin{itemize}
\item We leverage mutual information to measure hidden state transitions to identify redundant blocks, ensuring stable and reliable pruning.
\item We incorporate the Data Processing Inequality (DPI) to elucidate the relationship between the importance of entire contiguous blocks and that of individual blocks.

\item We develop the Fast-Block-Select algorithm, which iteratively updates block combinations to achieve a globally optimal solution while significantly improving the efficiency.
\end{itemize}

\section{Related Work}
\subsection{Block Pruning}
To design a pruning algorithm that is both simple and readily deployable for LLMs, several prior works have explored removing less critical blocks. For example, ShortGPT \cite{men2024shortgpt} estimates block importance via cosine similarity and applies greedy pruning strategies. Such greedy methods, however, often converge to local optima rather than discovering the globally optimal set of blocks to prune. Other approaches, like LaCo \cite{yang2024laco}, aim to compress the model by merging successive layers into their predecessors. This strategy, however, frequently falls short of the effectiveness achieved by directly removing layers. LLM-Streamline \cite{chen2024streamlining} prunes by consolidating blocks with the highest cosine similarity into a single block. While effective in some cases, excessive block merging can degrade model accuracy and require additional fine-tuning, increasing training overhead. Importantly, this method still relies on greedy selection. Similarly, approaches such as SLEB \cite{song2024sleb} and Shortened LLaMA \cite{kim2024shortened} perform iterative pruning based on layer importance. Yet, the repeated need to evaluate these metrics on a calibration set after each removal can lead to substantial computational costs.
\subsection{Mutual Information}
Mutual information is a fundamental metric for quantifying the dependency between two variables by measuring the information one variable provides about the other \cite{liu2022improving,nguyen2014effective,veyrat2009mutual}. Unlike conventional linear measures, it can capture non-linear relationships, making it a valuable tool for revealing complex data patterns and understanding intricate model dynamics \cite{vinh2012novel,pascoal2017theoretical}. Some studies achieve pruning by leveraging mutual information. MINT \cite{ganesh2021mint} leverages conditional geometric mutual information to induce sparsity by measuring the information dependency between adjacent layer filters. Some approaches \cite{huang2024large} adopt mutual information based estimators to identify redundant neurons for removal, while using carefully tuned estimators to guide the pruning process accurately. MIPP \cite{westphal2024mutual} is a structured, activation-based pruning method that preserves the mutual information between neurons across layers, enabling efficient and re-trainable pruning both before and after training. Additionally, other works \cite{fan2021layer} achieve a top-down, globally informed pruning strategy by retaining neurons with high mutual information.


\section{Method}
\subsection{Mutual Information Measures Block Importance}

During the inference phase of LLMs, the sequence outputs of the Transformer layers exhibit a high degree of similarity. This similarity primarily stems from a crucial design feature of the model: the use of residual connections. Specifically, the output of each layer is added to the output of the previous layer through these residual connections, thereby enabling the continuous transfer and accumulation of information across different layers of the model. Mathematically, the output of the \( (i + 1) \)-th Transformer layer can be represented as follows:
\begin{equation}
h_{i+1} = \text{Transformer}_{i+1} \left ( h_{i} \right ) + h_{i}
\end{equation}
Here, \( h_i \) and \( h_{i+1} \) denote the outputs of the \( i \)-th and \( (i + 1) \)-th layers, respectively, while \( \text{Transformer}_{i+1} \) represents the transformation function of the \( (i+1) \)-th layer. This additive mechanism ensures that the information from previous layers is retained and built upon, contributing to the overall robustness and depth of the model's representations.


Some studies identify less important blocks using cosine similarity \cite{men2024shortgpt}. However, the effectiveness of cosine similarity often decreases when dealing with non-linear relationships or complex, skewed data distributions, and it may not delve deeply into the specific interactions between variables. In contrast, {mutual information} \cite{liu2022improving,nguyen2014effective,veyrat2009mutual} presents a more robust and in-depth analytical approach. Unlike cosine similarity, which primarily focuses on the directionality of hidden states, mutual information delves into the actual information flow and dependency relationships between hidden states. This holistic perspective allows for a richer understanding of the underlying interactions within the model, thereby providing a more effective basis for optimization and deeper insights into model performance.

When the mutual information \cite{liu2022improving,nguyen2014effective,veyrat2009mutual} between the input and output states is unusually high, this typically indicates that the block's output is largely a direct reflection of its input state. This suggests that the block holds lower importance within the model. In other words, the output state does not significantly add new or important information but rather largely replicates the information from the input state, thereby introducing redundancy. In this case, the block may play a minor role in the overall functionality of the model. 

The transformation effect of the \(block_{i+1}\) is maximized when \( h_i \) and \( h_{i+1} \) are independent, and the mutual information is zero.
Conversely, the transformation function is minimal when \( h_{i+1} \) is completely determined by \( h_i \), such that for any \( h_i \) there exists a corresponding \( h_{i+1} \) for which \( P(h_{i+1} \mid h_i) = 1 \), and the mutual information is maximal.
Beyond these two extreme cases, as the transformation function exerts a greater effect, \( H(h_{i+1} \mid h_i) \) increases, while the mutual information \( I(h_i, h_{i+1}) \) decreases.

\begin{figure}
    \centering
    \includegraphics[width=1\linewidth]{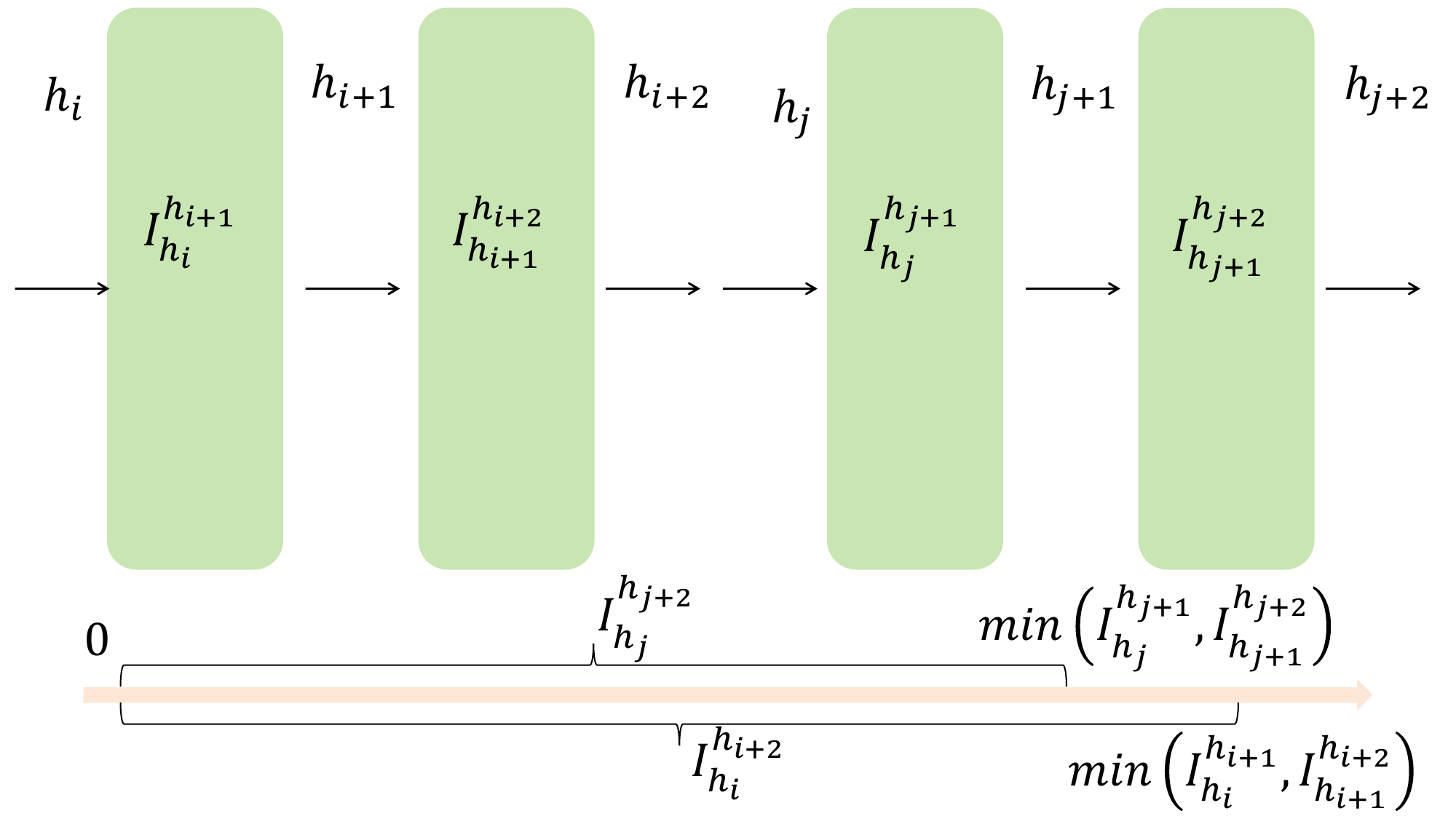}
    \caption{The relationship between the mutual information of the global continuous block and the local individual blocks.}
    \label{MI}
\end{figure}
\begin{figure*}
    \centering
    \includegraphics[width=1\linewidth]{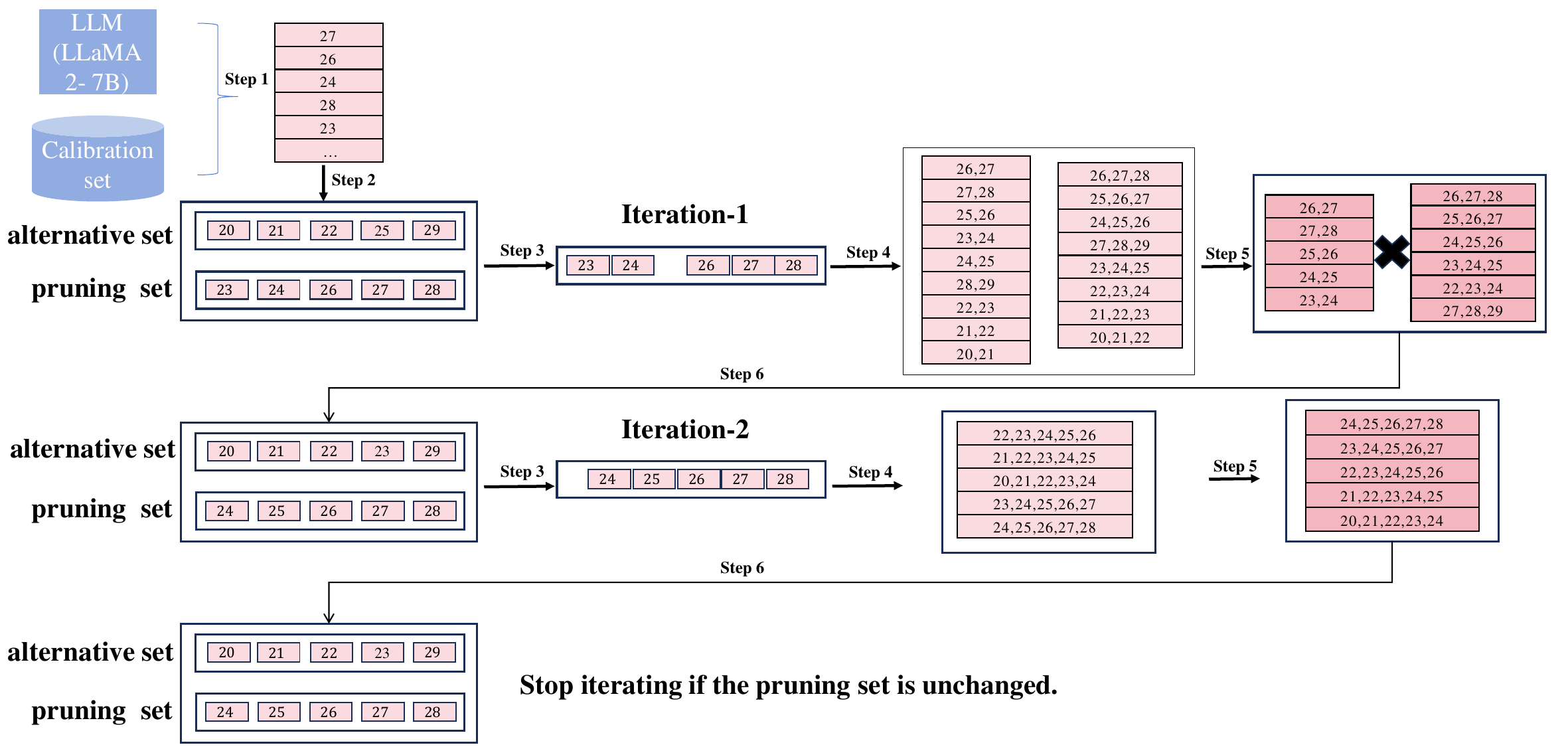}
    \caption{An overview of our method. The process of pruning 5 blocks in the LLaMA2-7B model. We provide a detailed description of the implementation for each step in Section \ref{Fast-Block-Select}.}
    \label{process}
\end{figure*}

\begin{algorithm}[!h]
\caption{Fast-Block-Select}
\label{algorithm:fast_block_select}

\KwIn{Model with $T$ blocks, calibration dataset $\mathcal{D}_{calib}$, target pruning number $N$, hyperparameter $k$}
\KwOut{Pruned block set $P$}

\BlankLine
\textbf{Single Block Importance Estimation:}

Execute the model on the calibration dataset $\mathcal{D}_{calib}$ to collect the input $b^{i}_{in}$ and output $b^{i}_{out}$ of each block $b^{i}$\;

\ForEach{$i \in \{1,2,\dots,T\}$}{
    $I^{i} \gets -MI(b^{i}_{in}, b^{i}_{out})$\;
}
Sort blocks by ascending $I^i$\;

\BlankLine
\textbf{Pruning \& Alternative Sets Initialization:}

$M \gets \min(N, T-N)$\;  
$P \gets$ top-$N$ least important blocks\;  
$A \gets$ top-$M$ least important blocks outside of $P$\;  

\BlankLine
$P' \gets \emptyset$\;

\textbf{Iterative Group Level Refinement:}

\Repeat{$P' = P$}{
    \If{$P' \neq \emptyset$}{
        $P \gets P'$\;
    }

    Decompose $P$ into contiguous groups $G = \{g_1, g_2, \dots\}$\;
    $R \gets \emptyset$\;

    \ForEach{$g \in G$}{
        $L \gets |g|$\;
        Form candidate set $C$ by generating all contiguous blocks of length $L$ from $P \cup A$\;

        \ForEach{$c=\{b^j,\dots,b^{j+L-1}\} \in C$}{
            $J_c \gets \sum_{k=j}^{j+L-1} I^k$\;
        }

        Sort $C$ by ascending $J_c$\;
        $K \gets \min(\lfloor \log L \rfloor + k, |C|)$\;
        $C^* \gets C_{1:K}$\;

        \ForEach{$c=\{b^j,\dots,b^{j+L-1}\} \in C^*$}{
            $I_c \gets -MI(b^j_{in}, b^{j+L-1}_{out})$\;
        }

        $R \gets R \cup C^*$\;
    }

    $P' \gets \textnormal{ConflictFreeSelect}(R)$\;
}

\Return{$P$}\;
\end{algorithm}

\subsection{Importance of Continuous Blocks}
In the previous section, we analyze the importance of blocks from the perspective of individual, isolated blocks. A potential issue arises: some individual blocks may appear unimportant on their own, but when considered in the context of the surrounding blocks as a whole, they can significantly contribute to the model performance. To gain a clearer understanding of this issue, we introduce the Data Processing Inequality (DPI) \cite{beaudry2011intuitive,merhav2012data,braverman2016communication}. The DPI states that when input \( h_i \) passes through a first-level block to obtain information \( h_{i+1} \), and then \( h_{i+1} \) is processed further by a second-level block to produce the output \( h_{i+2} \), the mutual information \cite{liu2022improving,nguyen2014effective,veyrat2009mutual} between \( h_i \) and \( h_{i+2} \) (\(I_{h_{i}}^{h_{i+2}}\)) satisfies the following relationship:
\begin{equation}
I_{h_{i}}^{h_{i+2}} \leq \min( I_{h_{i}}^{h_{i+1}}, I_{h_{i+1}}^{h_{i+2}} )
\end{equation}
As illustrated in Figure \ref{MI}, consider the combination of two blocks as an example. Suppose the importance of any block on the left ($h_i \rightarrow h_{i+1} \rightarrow h_{i+2}$) is lower than that of any block on the right ($h_j \rightarrow h_{j+1} \rightarrow h_{j+2}$). Under this scenario, the mutual information follows the following characteristics:
\begin{equation}
\min(I_{h_{i}}^{h_{i+1}}, I_{h_{i+1}}^{h_{i+2}}) \geq \max(I_{h_{j}}^{h_{j+1}}, I_{h_{j+1}}^{h_{j+2}})  
\end{equation}
Based on the DPI \cite{beaudry2011intuitive,merhav2012data,braverman2016communication}, the following formula can be derived:
\begin{equation}
    \overline{I_{h_{i}}^{h_{i+2}}} \geq \overline{I_{h_{j}}^{h_{j+2}}}
\end{equation}
Here, \(\overline{I_{h_{j}}^{h_{j+2}}}\) (\(\min(I_{h_{j}}^{h_{j+1}}, I_{h_{j+1}}^{h_{j+2}})\)) denotes the upper bound of \(I_{h_{j}}^{h_{j+2}}\), and the same applies to the \(\overline{I_{h_{i}}^{h_{i+2}}}\) (\(\min(I_{h_{i}}^{h_{i+1}}, I_{h_{i+1}}^{h_{i+2}})\)). The mutual information \(I_{h_{i}}^{h_{i+2}}\) is only guaranteed to have a higher upper bound than \(I_{h_{j}}^{h_{j+2}}\), but it is still possible that \(I_{h_{i}}^{h_{i+2}}\) is less than \(I_{h_{j}}^{h_{j+2}}\). Thus, a continuous block that follows two unimportant blocks is more likely to be unimportant as well. Nevertheless, the possibility that it could be important still exists. The same reasoning applies to the case of multiple continuous blocks, extending from the scenario of two continuous blocks.

To enhance the accuracy of block selection, we assess the importance of continuous blocks as a whole rather than individually. This approach prevents the unintended removal of blocks that, while seemingly less important on their own, significantly contribute to the overall performance of the model. For example, consider the LLaMA3.1-8B model, which consists of 32 blocks. When pruning 5 blocks, we need to evaluate the mutual information of 150 blocks, including both individual blocks and continuous block. We will elaborate on our detailed iterative block selection algorithm in Section \ref{Fast-Block-Select}.

\subsection{Fast-Block-Select (Iterative Updates)}
\label{Fast-Block-Select}

In practical applications, calculating the importance (measured by mutual information) of all contiguous blocks can be highly time consuming and resource intensive. For example, in the LLaMA2-7B model, there are as many as \(528\) possible combinations of contiguous blocks, making the task extremely complex. Therefore, it is crucial to explore methods that can accelerate the block selection process. A straightforward idea might be to use dynamic programming to tackle this problem. However, since the importance of contiguous blocks does not have a strict quantitative relationship with the importance of their sub-blocks, this approach is not viable. To address this challenge, we have designed a heuristic block selection method called {Fast-Block-Select} to significantly enhance the speed of block selection. Assuming the model consists of a total of \(T\) blocks, we aim to prune \(N\) blocks. The process can be broken down into the following steps (Figure~\ref{process} provides a detailed illustration of our process for pruning \(5\) blocks in the LLaMA2-7B model):

\textbf{Step 1 : }We employ a calibration set to obtain the importance of each independent block within the model and subsequently rank them in ascending order of importance. In Figure \ref{process}, we employ a calibration set to perform inference on LLaMA2-7B and obtain a list of blocks sorted in ascending order of importance (27, 26, 24, 28, 23 ...).

\textbf{Step 2 : }We select the top \(N\) blocks according to their importance ranking to form the pruning set. Additionally, we identify the \(M\) least important blocks that are outside the pruning set to form the alternative set, where \(M\) satisfies the following conditions:
\begin{equation}
M = \min({N}, {T} - N)
\end{equation}
The number of elements in the alternative set ($M$), is typically set to match the number of elements in the pruning set ($N$), and should not exceed the total number of remaining blocks (\({T} - N\)). In Figure  \ref{process}, both \( N \) and \( M \) are set to 5. The pruning set is defined as \( \{23, 24, 26, 27, 28\} \), and the alternative set is defined as \( \{20, 21, 22, 25, 29\} \).

\textbf{Step 3 : }We categorize the blocks in the pruning set into groups to form contiguous block sets. In Figure \ref{process}, in {Iteration-1}, we group the pruning set into \( \{23, 24\} \) and \( \{26, 27, 28\} \). In {Iteration-2}, since all elements in the pruning set are contiguous, there is only one group: \( \{24, 25, 26, 27, 28\} \).

\textbf{Step 4 : }For each contiguous block within the contiguous blocks set, we generate all possible contiguous block sets of matching lengths by utilizing blocks from both the pruning set and the alternative set. We assess the importance of each contiguous block by summing the importance values of its constituent individual blocks, and subsequently sort these contiguous blocks in ascending order of their estimated importance. This method provides a reasonable approximation of the importance of contiguous blocks, as a sequence composed of multiple important individual blocks is more likely to be significant. Importantly, this approach eliminates the need to directly compute the mutual information between the inputs and outputs of the contiguous blocks; instead, it simply involves aggregating the mutual information of the individual blocks. In Figure \ref{process}, during Iteration-1 (with contiguous block lengths of 2 and 3), we construct possible contiguous block sets using elements from the pruning set and the candidate set. We also estimate the importance of these contiguous blocks and sort them, resulting in the sets $\{\{26, 27\}, \{27, 28\}, \{25, 26\} \ldots\}$ for the set $\{23, 34\}$. For the set $\{26, 27, 28\}$, we construct the contiguous block sets $\{\{26, 27, 28\}, \{25, 26, 27\} \ldots\}$. In Iteration-2 (with contiguous block length of 5), we construct the contiguous block sets $\{\{22, 23, 24, 25, 26\} \ldots\}$.

\textbf{Step 5 : }For each group, we compute the exact mutual information of the top \( K \) contiguous blocks, and subsequently rank these blocks based on their mutual information. Here, \( K \) is determined by the following conditions:
\begin{equation}
K = \min(\lfloor \log L \rfloor + k, l)
\end{equation}
Here, \( L \) denotes the length of the corresponding contiguous block, while \( l \) represents the number of elements in the continuous block set. Additionally, \( k \) signifies the number of extra elements we consider (a hyperparameter). As \( L \) increases, the error in measuring the importance of a continuous block based solely on the sum of the importance of individual blocks also increases. Therefore, we incorporate \( \log L \) into the calculation of \( K \), ensuring that \( K \) increases with \( L \). Additionally, the upper bound of \( K \) cannot exceed \( l \). In Figure \ref{process}, in Iteration-1, \( L \) is 2 and 3, while \( K \) is 5 and 6, respectively. In Iteration-2, \( L \) is 5, and \( K \) is 5.

\textbf{Step 6 : }For each group of contiguous blocks, we select the combination with the highest sum of mutual information that does not conflict as the updated block combination for this iteration. In Figure \ref{process}, in Iteration-1, we select \(\{24, 25\}\) and \(\{26, 27, 28\}\), resulting in a new pruning set \(\{24, 25, 26, 27, 28\}\). In Iteration-2, we select \(\{24, 25, 26, 27, 28\}\). The pruning set changes between before and after Iteration-1, so the iterative algorithm continues. However, the pruning set remains unchanged between before and after Iteration-2, so the iteration stops.

The steps 3, 4, 5, and 6 are executed in each iteration. This process continues until the selected blocks remain unchanged from one iteration to the next. Our method takes into account the influence of blocks more comprehensively and thoroughly both before and after each iteration, which enables it to obtain either better or equivalent solutions (never worse). This provides a significant guarantee of convergence. Extensive testing on a wide range of models and data also show that our method does not exhibit oscillation phenomena, further demonstrating its effectiveness.

\section{Experiments}
\begin{table*}[]

\resizebox{\textwidth}{!}{
\begin{tabular}{c|c|c|ccccccccccc}
\hline
\hline
\textbf{Models}             & \textbf{Methods} & \textbf{Ratio} & \textbf{Winogrande} & \textbf{PIQA}  & \textbf{WSC}   & \textbf{WNLI}  & \textbf{SST-2} & \textbf{RTE}   & \textbf{QNLI}  & \textbf{CB}    & \textbf{ARC-e} & \textbf{ARC-c} & \textbf{Avg.}  \\ \hline
\multirow{5}{*}{Llama2-7B}  & Dense            & 0.00\%         & 72.85               & 78.24          & 54.81          & 59.15          & 85.89          & 66.06          & 54.6           & 60.71          & 79             & 47.78          & 65.91          \\
                            & SliceGPT         & 15.34\%        & 63.06               & 67.03          & \textbf{63.46} & 45.07          & 52.64          & 59.57          & 50.63          & 42.86          & 55.26          & 34.56          & 53.41          \\
                            & LLM-Pruner       & 15.30\%        & 64.72               & \textbf{76.06} & 36.54          & 56.34          & 65.83          & 52.71          & \textbf{51.20} & 46.43          & 64.86          & 36.69          & 55.14          \\
                            & ShortGPT         & 15.32\%        & 67.72               & 71.6           & 36.54          & 43.66          & 49.2           & 53.43          & 50.19          & 39.29          & 65.66          & \textbf{39.42} & 51.67          \\
                            & MI-PRUN          & 15.32\%        & \textbf{69.69}      & 73.5           & \textbf{63.46} & \textbf{60.56} & \textbf{83.94} & \textbf{60.29} & 50.91          & \textbf{62.5}  & \textbf{67.21} & \textbf{39.42} & \textbf{63.15} \\ \hline
\multirow{5}{*}{Llama2-13B} & Dense            & 0.00\%         & 75.61               & 79.71          & 53.85          & 66.2           & 87.61          & 69.31          & 58.56          & 80.36          & 81.82          & 53.16          & 70.62          \\
                            & SliceGPT         & 25.28\%        & 70.72               & 63.17          & 44.23          & 43.66          & 51.15          & 52.71          & 50.58          & 41.07          & 65.03          & 33.79          & 51.61          \\
                            & LLM-Pruner       & 25.35\%        & \textbf{71.74}      & \textbf{73.07} & 36.54          & 43.66          & 66.86          & 52.71          & 49.90          & 42.86          & \textbf{66.08} & 32.42          & 53.58          \\
                            & ShortGPT         & 24.37\%               & 64.17               & 71.93          & 58.65          & 59.15          & 50.57          & 68.95          & 49.83          & 57.14          & 50.17          & \textbf{40.53} & 57.11          \\
                            & MI-PRUN          & 24.37\%              & 59.59               & 72.36          & \textbf{59.62} & \textbf{66.2}  & \textbf{70.87} & \textbf{69.31} & \textbf{50.61} & \textbf{78.57} & 60.02          & 40.27          & \textbf{62.74} \\ \hline
\multirow{3}{*}{Qwen-7B}    & Dense            & 0.00\%         & 71.98               & 79.22          & 73.08          & 64.79          & 94.15          & 83.75          & 74.24          & 76.79          & 79.04          & 48.04          & 74.51          \\
                            & ShortGPT         & 13.11\%               & 67.4                & 73.01          & \textbf{63.46} & \textbf{63.38} & 93.12          & \textbf{80.87} & \textbf{50.54} & 60.71          & 62.96          & 35.84          & 65.13          \\
                            & MI-PRUN          & 13.11\%              & \textbf{69.3}       & \textbf{73.78} & \textbf{63.46} & 61.97          & \textbf{94.04} & 71.84          & \textbf{50.54} & \textbf{62.5}  & \textbf{66.25} & \textbf{39.93} & \textbf{65.36} \\ \hline
\multirow{3}{*}{Qwen-14B}   & Dense            & 0.00\%         & 75.3                & 80.01          & 69.23          & 73.24          & 95.07          & 80.51          & 74.81          & 87.5           & 81.14          & 52.47          & 76.93          \\
                            & ShortGPT         & 15.58\%               & 67.4                & 70.18          & 36.54          & \textbf{71.83} & 94.15          & 80.51          & 49.46          & \textbf{87.5}  & 63.72          & 39.33          & 66.06          \\
                            & MI-PRUN          & 15.58\%              & \textbf{69.46}      & \textbf{72.25} & \textbf{81.73} & \textbf{71.83} & \textbf{95.41} & \textbf{80.87} & \textbf{69.89} & 83.93          & \textbf{67.38} & \textbf{43.69} & \textbf{73.64} \\ \hline \hline
\end{tabular}
}
\caption{Comparison of pruning methods on multiple natural language benchmarks.}
\label{Comparison}
\end{table*}

\subsection{Experimental Setup}
\begin{figure}
    \centering
    \includegraphics[width=1\linewidth]{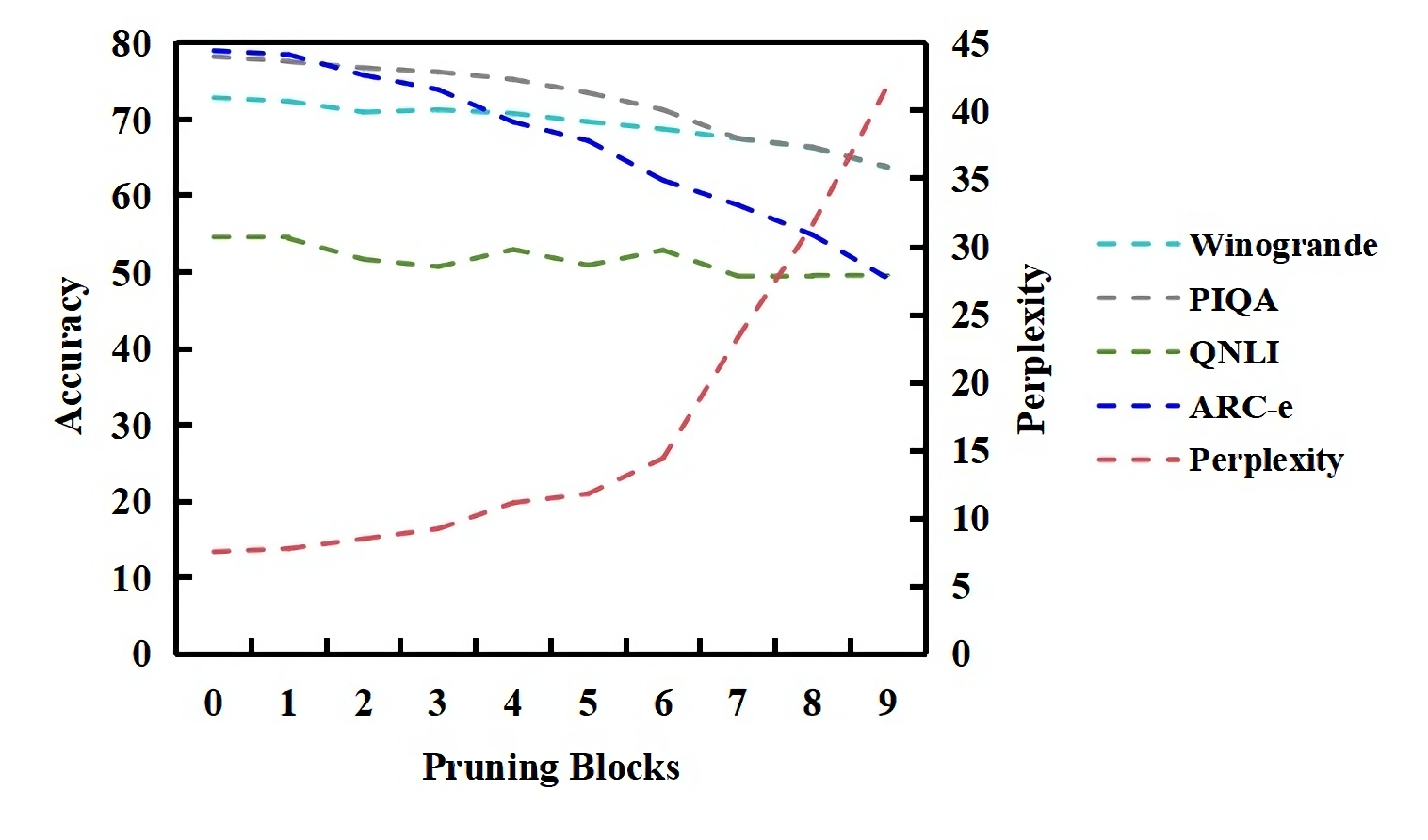}
    \caption{\centering{Performance of  Llama2-7B with increasing pruning blocks.}}
    \label{increasing}
\end{figure}
\begin{figure}
    \centering
    \includegraphics[width=1\linewidth]{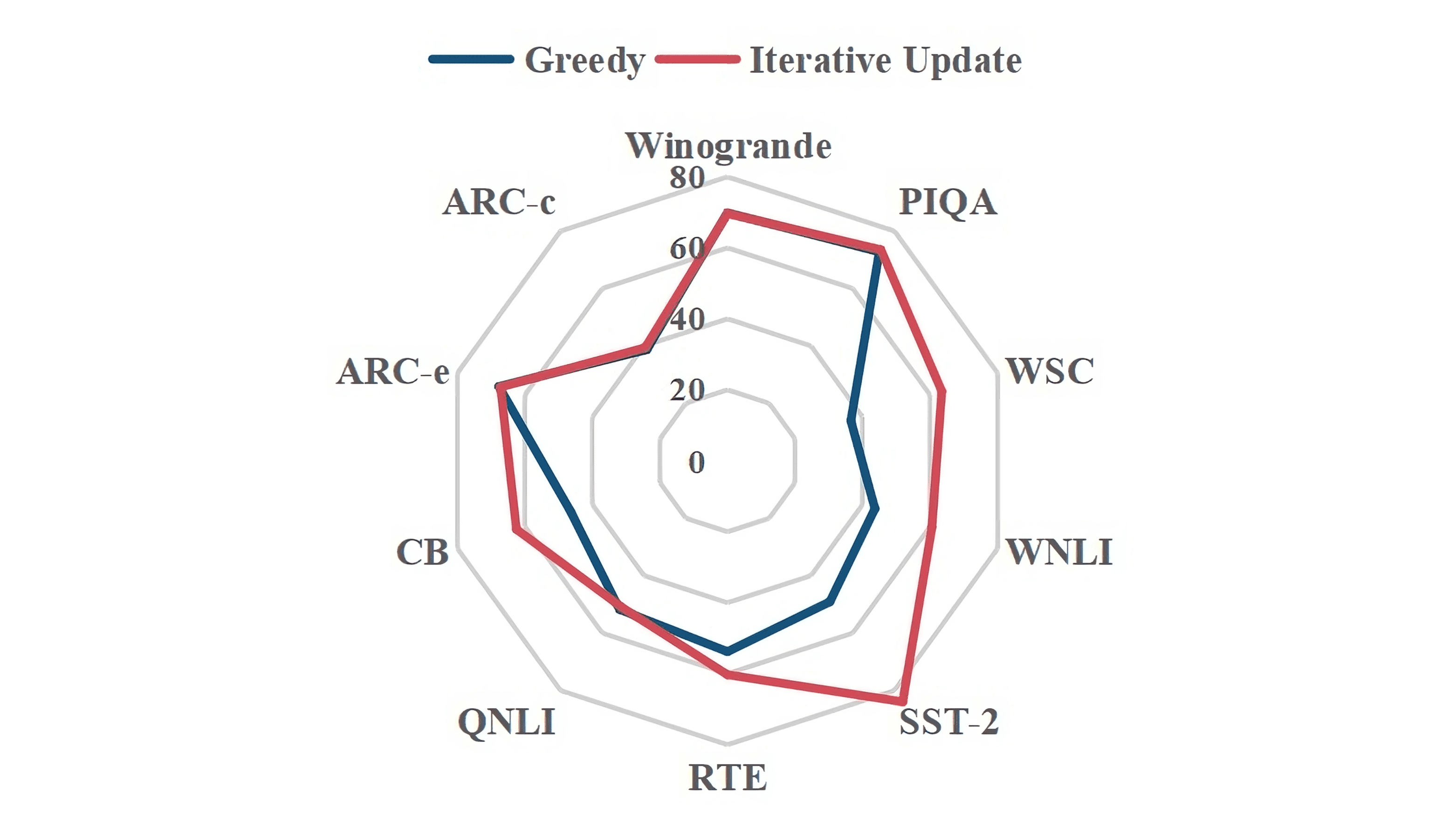}
    \caption{\centering{Performance comparison between Greedy Strategy and Iterative Update.}}
    \label{Greedy}
\end{figure}
\textbf{Models.} To showcase the effectiveness of our method, we execute a series of experiments utilizing several prevalent open-source language models, encompassing Llama2-7B \cite{touvron2023llama}, Llama2-13B, Qwen-7B \cite{qwen} and Qwen-14B. These models are all expansive, grounded in the Transformer architecture that operates solely with a decoder. The Llama2 models are trained across an extensive dataset exceeding two trillion tokens. Qwen-series are pre-trained using a dataset that totals 3 terabytes, covering a wide range of multilingual content including public web documents, encyclopedias, books, and code, with a primary focus on Chinese and English.

\textbf{Benchmarks. }In order to comprehensively assess the impact of pruning on the capabilities of large language models, we conduct an evaluation using widely prevalent benchmarks. Our principal benchmarks include  Winogrande , PIQA \cite{Bisk2020}, WSC \cite{kocijan2020review}, WNLI, SST-2 \cite{socher-etal-2013-recursive}, RTE \cite{poliak2020survey}, QNLI \cite{wang2018glue}, CB \cite{talmor-etal-2019-commonsenseqa}, ARC-e \cite{yadav2019quick} and ARC-c \cite{yadav2019quick}. These datasets serve as critical benchmarks for evaluating the performance of language models across a spectrum of NLP tasks, including question answering, text entailment, sentiment analysis, and commonsense reasoning. Additionally, we supplement our evaluation with a perplexity analysis on the C4 \cite{raffel2020exploring} dataset.

\textbf{Baselines.} To evaluate the effectiveness of our method, we compare the following pruning methods for
large language models:
\begin{itemize}
\item \textbf{LLM-Pruner: }The method adopts structural pruning that selectively removes non-critical coupled structures based on gradient information, maximally preserving the majority of the LLM’s functionality \cite{ma2023llm}. 
\item \textbf{SliceGPT: }SliceGPT is a new post-training sparsification scheme which replaces each weight matrix with a smaller 
 matrix, reducing the embedding dimension of the network \cite{ashkboos2024slicegpt}.
\item \textbf{ShortGPT: }It is a straightforward pruning approach: layer removal, in which we directly delete the redundant layers in LLMs based on their BI scores \cite{men2024shortgpt}. 
\end{itemize}

\textbf{Implementation Details.} During the pruning process,
if the product of the total number of transformer blocks in a model and the target sparsity is not an integer, we round up the product to determine the number of transformer blocks to remove.
 We conduct experiments using two different calibration sets: the WikiText-2 training dataset \cite{merity2016pointer} and the Alpaca training dataset \cite{taori2023stanford},  varying the calibration set size and sequence length. Additionally, when comparing the performance of different methods, we employ the same experimental settings, including the calibration set and pruning rate.

\subsection{Main Results}

To validate the effectiveness of our proposed method, comparative experiments are conducted on Llama2 and Qwen, employing standard benchmarks and baselines commonly utilized in the assessment of large language models. The experimental results
are shown in table \ref{Comparison}W. The results suggest that the models pruned via our proposed method have demonstrated enhanced overall performance when compared with the baseline methods, maintaining most of the large language model’s capabilities. We also conduct ablation studies on different pruning ratios as well as on the iterative update mechanism.



\section{Conclusion}
In this paper, we propose MI-PRUN, a mutual information–based pruning framework for large language models (LLMs). Specifically, our method leverages mutual information to quantify the dependency between consecutive hidden states, enabling the identification of redundant transformer blocks that contribute little to information propagation. To further enhance block-level importance estimation, we incorporate the Data Processing Inequality (DPI), which establishes a principled connection between the importance of contiguous block segments and that of individual blocks. Building on this formulation, we introduce Fast-Block-Select, an efficient algorithm that iteratively refines block combinations while substantially reducing the computational cost of block selection. As a result, MI-PRUN achieves significant inference acceleration with minimal performance degradation, making it a practical and effective solution for deploying LLMs under resource-constrained scenarios.


\bibliography{custom}

\end{document}